# Incremental Probabilistic Inference

Bruce D'Ambrosio
Department of Computer Science
Oregon State University
dambrosi@research.cs.orst.edu

## Abstract

Propositional representation services such as truth maintenance systems offer powerful support for incremental, interleaved, problem-model construction and evaluation. Probabilistic inference systems, in contrast, have lagged behind in supporting this incrementality typically demanded by problem-solvers. The problem, we argue, is that the basic task of probabilistic inference is typically formulated at too large a grain-size. We show how a system built around a smaller grain-size inference task can have the desired incrementality and serve as the basis for a low-level (propositional) probabilistic representation service.

## 1 Introduction

Propositional representation services such as truth maintenance systems offer powerful support for incremental, interleaved, problem-model construction and evaluation[1]. However, while these systems provide strong facilities for exploring alternate problem formulations, they provide little control over tradeoffs between inferential completeness and complexity, and limited facilities for ranking alternate solutions. In theory, probabilistic representations are ideal for reasoning about tradeoffs, but existing probabilistic inference systems are intended for inference within static models, are inefficient, and few offer control over resource consumption. In summary, no existing general-purpose low-level (propositional) representation service provides incrementality with respect to model revision and resource usage in a theoretically sound manner. In this paper we begin by offering a redefinition of the basic probabilistic inference task. We sketch how an inference engine which performs this task can serve as the core of an incremental probabilistic representation service, and report on progress to date in actually constructing such a system.

A belief net is a compact representation for the joint probability distribution over a set of variables. The representation consists of a directed acyclic graph over the variables and a set of marginal and conditional probability distributions, one for each variable [23]. While probabilistic inference in general is NP-hard [2], current state-of-the-art belief-net algorithms exploit the independence information in the graph to construct efficient computations for probability distributions not explicitly stored in the belief net [23], [17], [25]. However, in practice computational cost still grows rapidly [18] (except in the case of a few special-case net topologies), limiting application of these techniques to belief nets with a few hundred variables at most.

Also, the services offered by current belief-net based systems are not well matched to the needs of higher level problem solvers. As we discussed in [3] and [4], problem solvers typically interleave model construction, revision, and evaluation. One class of propositional representation service, truth maintenance systems [11], [8], [20], [7], is optimized for this kind of use: truth maintenance systems typically provide incremental (but monotonic) model construction facilities, and incrementally update inference when the propositional model is expanded. Also, while resource incrementality was not a feature of early TMS's, deKleer has found it desirable to extend the ATMS to include resource incrementality through various "focusing" mechanisms [10], [9].

## 2 Desiderata

We believe a low-level representation service should have two key properties: it should be *Incremental* and *Efficient*. A system is *Incremental* with respect to some capability to the extent that it can make use of the results of previous computations to reduce the cost or improve the quality of results for subsequent computations.

[1] Acknowledgment: This work supported by NSF 91-00530, AFOSR, and the Oregon Advanced Computing Institute.

For example, a system would be incremental with respect to queries if it took advantage of results computed during processing of earlier queries in the processing of some subsequent query. We identify four aspects of incrementality possible in probabilistic inference:

1. Resource incrementality: Any practically usable system must offer facilities for computing approximate responses to queries. Incrementality with respect to resources enables a system to use increments of time to refine estimates or bounds. This give the problem solver control over the time/quality tradeoff in inference.

2. Query incrementality: Many probabilistic inference systems automatically compute the answer to a fixed set of queries (eg, the set of marginal probabilities for all the nodes in the net), and most have no capability to process queries outside this set. Incrementality with respect to queries enables a system to accept multiple queries, and to use partial results computed during processing of earlier queries to simplify processing of subsequent queries.

3. Evidence incrementality: Evidence typically arrives over time: A robot turns to scan a new part of the scene, a medical lab reports a new test result, and so on. Incrementality with respect to evidence enables a system to update its internal representations when new evidence arrives, rather than recompute all queries from the initial belief net. Most modern belief net algorithms possess evidence incrementality.

4. Representation incrementality: A belief net is an impoverished representation: it is a minor extension of a propositional logic. We believe, therefore, that resource incrementality within a static belief net is not enough, but rather that inference within a partial problem representation must be able to be interleaved with representation extension operations, so that a problem solver can heuristically search towards an appropriate problem representation. Incrementality with respect to representation extension enables a system to reuse results from prior computations even when the representation on which those computations is based is modified between queries.

The last form of incrementality stated above may seem a bit extreme. Yet, Wimp [1], a problem solver of the kind sketched above, suffered severly because the belief net service it used was not incremental with respect to representation extensions.

**Efficiency** The goal of incrementality is efficiency. Not all efficiency concerns, however, are captured under the rubric of incrementality. A representation service is *Efficient* to the extent that it maximizes the information gain with respect to a query per resource increment. Again, we can identify several desirable forms of efficiency:

1. Efficiency with respect to network structure: There are three kinds of structure which can be exploited: the network topology, intra-distribution qualitative structure, and quantitative structure. All modern belief net algorithms exploit the conditional independence information contained in the topology of a belief net to reduce computational complexity. However, there is often considerable structure within the conditional distributions in a belief net [5], [13], [26], [12]. This structure can and should be exploited to improve efficiency. For a discussion of how this structural information is captured and exploited in SPI see [5]. Finally, there is often considerable numeric structure within a belief net, in the form of skewness of distributions (a distribution is *skewed* when one of the probability masses in the distribution is larger than the others, we will formalize this later). Several systems have explored exploitation of this structure [22], [16], [15].

2. Efficiency with respect to resource incrementality: We expect an incremental system to be only minimally more expensive than a non-incremental system on comparable tasks.

## 3  Term Computation - a new task definition

Probabilistic inference in belief nets, as currently defined, is generally taken to be the computation of a predefined set of prior or posterior marginal, conjunctive, or conditional probability distributions in a fixed network. This is often an unnecessarily restrictive formulation of the problem. The actual computation of any prior or posterior probability can in general be viewed as a sum over a number of terms (in the extreme case, this occurs as marginalization of the full joint). While the number of terms to be computed is exponential in the size of the network, the time complexity of computation of a single term is linear in the number of nodes relevant to the query. Consider the network shown in figure 1.

In this network:

$P(C_t) = \sum_{A,B} P(C_t|B,A) * P(B) * P(A)$
$= .95 * .9 * .75 + .95 * .1 * .2 + .05 * .85 * .2 + .05 * .15 * .75$
$= .64125 + .019 + .0085 + .005625$

We take the computation of a single term as an appropriate primitive task for probabilistic inference, and next show how an inference system with the needed incrementality properties can be built around it.

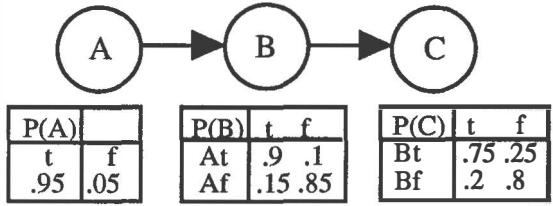

Figure 1: Simple Belief Net

A term computation approach will be interesting only if we can get a significant amount of information through the computation of a small number of terms. While there are many ways this might arise[2], we motivate the approach through the introduction of a critical assumption: we assume that most distributions in a belief net are "skewed."

**Definition 1** *A marginal probability distribution is skewed if one mass element is at least $(n-1)/n$, where n is the number of nodes in the network. A conditional distribution is skewed if each row has this property. In this case it need not be the same element in each row.*

If all the distributions in a belief net are skewed, then most of the probability mass for many queries is contained in the largest few terms[3]:

**Theorem 1** *Given a Belief-net over n two-valued variables such that all distributions are skewed with a larger mass of at least $(n-1)/n$, then the $n+1$ largest terms in the joint distribution across the variables contain a total mass of greater than $2/e$.*

Note that this result is not based on any assumptions about the structure of the network. The degree of skewness assumed in the above theorem may seem extreme. However, it is quite natural in many applications, such as failure modeling of engineered systems. Thus, our answer to the question of which terms to compute will be to compute the largest terms first.

It would be easy to construct a term computation system which merely enumerated elements of the full joint distribution across all variables in a network, as in our example. Indeed, some existing proposals for anytime probabilistic inference essentially do this [15]. However, such an approach can be inefficient. There are several sources for this inefficiency: First, there would be a time inefficiency due to unnecessary repetition of sub-computations (eg, the computation of $P(B_t|A_t) * P(A_t)$ in our example). Second, there would be space inefficiency resulting from keeping each term separate. Finally, it is not obvious how such simple methods can be made incremental with respect to newly arriving evidence, queries, or belief net extensions.

---
[2]For example, through domain dependent knowledge of paradigmatic "cases".

[3]Proof in longer report.

### 3.1 Basics of Probabilistic Inference

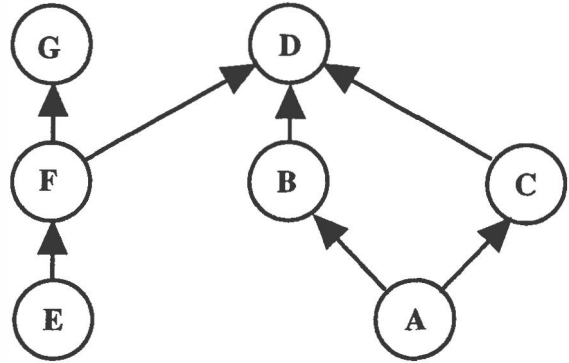

Figure 2: Paradigmatic Belief Net

Developments in exploiting the probabilistic independence relations expressed in the topology of a belief net provide the necessary basis for designing computations which address these problems. In general, the sparser a belief net, the more finely any computation can be partitioned into independent sub-computations which share only a small number of variables. For example, given the net in figure 2, a query for $P(D)$ can be computed by first computing the full joint probability distribution, then marginalizing over all variables except $D$:

$P(D) = \sum_{A,B,C,E,F} P(D|B,C,F) * P(F|E) * P(E) * P(B|A) * P(C|A) * P(A)$

However, a much more efficient form of the computation is:

$P(D) = \sum_F (\sum_E P(F|E) * P(E))$
$\quad * (\sum_{B,C} P(D|B,C)$
$\quad * (\sum_A P(B|A) * P(C|A) * P(A)))$

Having done this, we can eliminate redundant computation by caching intermediate results. Similarly, we can reduce the space requirement by combining terms when their bindings differ only on variables not needed in the remainder of the computation. In the extreme, each of these can reduce the corresponding complexity (time and space) for computing each term beyond the first from $n$ to $Log(n)$, where $n$ is the number of variables relevant to a query.

Construction of an optimal evaluation poly-tree for an arbitrary query set is a hard problem [19]. However, simple, polynomial-time greedy heuristics perform quite well, and are described in [19]. This previous work was performed in the context of exact query evaluation (that is, computation of all terms), but the theory remains applicable, and so will not be repeated in detail here. The basic constraint is that a variable may not be marginalized out unless it appears only in the subtree below the node at which the marginalization is to take place. One constraint we add for term computation is that evaluation polytrees are built such that, when searched depth-first left-to-right, the marginal distribution for a root variable will be encountered before any conditional distributions naming the variable. We enforce this constraint by constructing the polytree bottom up, starting from the belief net roots. The following is a sketch of the algorithm we currently use to build the tree for a single query:

- Select the nodes relevant to the query using a d-separation algorithm.

- Divide the nodes into layers, according to distance from the furthest ancestor root.

- For each layer, starting from the roots:
  - Partition the layer and the factors from the previous layer into independent factors (factors with no overlapping variables).
  - Label each new factor with the variables it contains which are needed by nodes in descendant layers.
  - Build an internal evaluation tree for each factor using a modified version of the set factoring algorithm of Li, which always orders children of a eval tree node so that a marginal, if present, is on the left.

Consider the net in figure 2, and assume our queries are for $G$ and $D$. The expression for $P(D)$ has been given earlier. The expression for $P(G)$ is:

$$P(G) = \sum_F P(G|F) * (\sum_E P(F|E) * P(E))$$

We can efficiently combine these two expressions into a single evaluation poly-tree, as shown in figure 3

In the following section we first develop the basics of term computation (which is inherently incremental with respect to resource consumption, although not efficient, as we shall see) for a static network, set of evidence, and set of queries. We then describe how the fundamental computation can be made efficient and incremental with respect to queries, evidence, and net extension.

### 3.2 Basics of term computation

The elementary primitive out of which we build a term computation system is the construction of a stream of terms for some node in the evaluation poly-tree for a set of queries. A stream of terms is a closure (a function with all of its parameters bound to some node in the evaluation poly tree) which, each time it is invoked, returns the next term for that node. This stream will be constructed, recursively, by combining streams of terms from child nodes in the poly-tree. We first describe the evaluation poly-tree and its construction, then explain the term computation process.

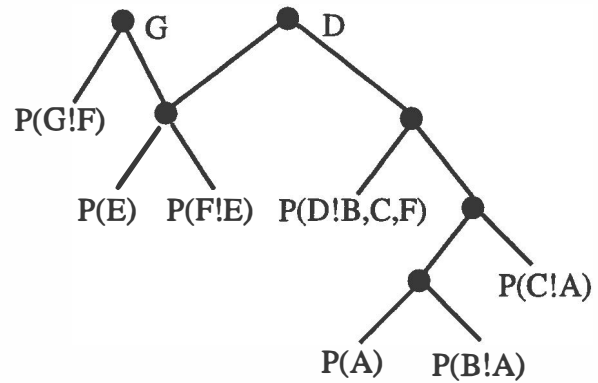

Figure 3: Evaluation Poly-tree for sample query set

### 3.3 Term computation

Given an evaluation poly-tree for a query set, we can recursively define a primitive operation at each node in the tree: generation of the next term at that node. Term generation is simple: each term is generated by forming the product of a term from the left child and a term from the right child. There are, however, two issues to consider: (1) Control: the decision of which term to compute next; (2) Efficiency: Basic term computation as outlined is quite inefficient. We will show how it can be made efficient.

**Control** We earlier stated that we would attempt to minimize the number of terms computed by computing largest terms first. We are exploring both optimal (A*[4]) and simple greedy search methods. A* requires two measures, a measure of "distance so far" and a heuristic estimate of remaining distance. We use the mass computed so far as the inverted "distance traveled so far," and the partial value returned by a partial subterm as our heuristic estimate. This is an admissible heuristic, and so guarantees that the largest term will be in front of the agenda upon termination[5]. Prob-lem solver guidance can be provided in the form of a "scaling function" which has access to term bindings and can scale the probability masses before they are used to order the search agenda.

---

[4] Actually, Z*, since step costs are multiplicative.

[5] This selection criterion is similar to the techniques used by deKleer [10] and Henrion [15]. Both use search on restricted classes of networks for the diagnostic task of finding most likely composite hypotheses, with good results. One contribution of our work is to show how this technique can be used in a more general setting.

**Efficiency** As we discussed earlier, a naive enumeration of all terms is inefficient in use of both space and time. The space inefficiency arises because the number of terms computed in response to any query is exponential in the number of relevant variables. However, the major advance offered by recent developments in probabilistic inference is reduction of the exponent for computation of complete distributions from number of relevant variables to number of relevant variables manipulated at once at any node in the evaluation polytree. We should not have to pay a higher price simply to achieve incrementality. We can achieve this efficiency by merging, at each node, completed terms which are distinguished only by bindings on variables not needed at higher levels of the evaluation poly-tree. This creates two problems. First, a term which has already been incorporated into streams at higher levels in the evaluation poly-tree can suddenly have its value change (positively). Simple dependency tracking mechanisms suffice to record the information needed to update these higher terms. Second, exactly what does the Z* guarantee now mean? In poly-tree nodes where marginalization takes place, a partial term can be extended in two ways: by multiplying its value by terms from remaining distributions, or by adding additional ground terms[6]. While we use a heuristic which is admissable in its estimate of the effect of the former, our heuristic is inadmissable with regard to the latter (because it ignores marginalization). This means we can only make a relatively weak statement about terms in streams generated from poly-trees containing marginalization: that the first term returned will be that term whose lower bound is highest after considering all complete ground subterms computed so far. Note that the term need not be "complete" in the sense that further ground terms may be added into it during later computation. It is, however, complete in the sense that it is a sum of a set of complete ground terms.

SImilarly, the basic method is quite time inefficient. This is because in the course of search a node will be typically be expanded many times. Marginalization removes some, but not all, of this redundancy. Caching streams, indexed by the node and the relevant bindings, removes the remaining redundancy, and makes term computation as space efficient as standard exact algorithms when computing all terms.

**Complexity** The key assumptions we make are that: (1) the probability distributions are sufficiently skewed and; (2) the graphical structure of the belief net is sufficiently sparse. Under these assumptions, the evaluation poly-tree will be such that the total number of terms computed in all streams, in the course of computing the first $n$ term requests for each query in the query set, will be $n$ times the number of nodes in the poly-tree. Since the poly-tree is a binary tree, this in turn is $2n$ in the number of variables relevant to the query set. All the operations we have described are either constant time, linear, or at worst $nlog(n)$ (reordering the agendas) in the number of terms in an agenda. Therefore, the total complexity, in the admittedly most optimistic case, is $2n^2 log(n)$ where $n$ is the number of variables relevant to a query set and the number of terms requested. Our experience in actually applying this procedure to three tasks, computation of marginal probabilities, most likely composite hypotheses, and complete decision analysis, confirms that this estimate is in fact realistic for a typical class of belief nets describing decision models for diagnosis and control of simple digital circuits. The biggest unknown in all of this is, of course, search complexity. We present some experimental data on this point later.

## 4 Error Estimates

Under the skewness assumption, the total mass contained in the largest $m$ terms (ignoring marginalization) from a computation involving $n$ variables is at least:

$$\int_0^Q BinomialDistribution[n, 1/n]$$

Where:

$$m = \Sigma_0^Q Binomial[n, i]$$

This later is difficult to solve for $Q$. For small $m$ ($Q < 3$), we can approximate it as:

$$Q + 2Log_n[Q - 1] = Log_n m$$

Using a normal to approximate the binomial, we can approximate the total mass as (the "+.5" adjustment to $Q$ in the cumulative makes the estimate more accurate for small values of $Q$):

$$\int_{-\infty}^{Q+.5} NormalDistribution[1, ((n-1)/n)^{\frac{1}{2}}]$$

In most real nets, some distributions will not meet skewness requirements, while others will be more skewed than required (eg, deterministic). We can use least-squares fit of the cumulative density to obtain an effective $n$ in these cases, and so estimate convergence rate, remaining mass, and normalization factor when needed.

---

[6] A "ground" term is one with a unique binding for each variable in the subtree rooted by the poly-tree node under consideration.

## 5  Making Term Computation Incremental

The basic process sketched above is incremental and efficient with respect to computation of additional terms for a static query set. In this section we discuss extensions to the basic method to make it incremental with respect to new queries, evidence, and model extensions/revisions.

**Queries**  Given the recursive query decomposition process we sketched above, it should be obvious that the process is inherently incremental with respect to newly arriving queries. One can incrementally elaborate the evaluation tree for the new query, top-down, testing for existence of a stream for a subquery before creating a new stream.

**Observations**  New evidence (in the form of assertions that a variable has been observed to take on a specific value) affects an existing term computation structure in several ways: (1) Terms which are bound to unobserved values of the evidence variable must be removed from all streams in which they appear; (2) Terms for consequents of the observed variable are no longer dependent on the antecedents of the evidence variable, requiring pruning of the mass dependency structure of the affected terms and propagation of the resulting mass changes upward through the evaluation poly-tree. (3) Certain query evaluation subtrees will require additional child subtrees (effectively, conditioning on the new evidence - see the discussion of d-separation in, for example, [23] for further details). We handle this by invalidating and recomputing all streams on a line from the poly-tree node at which a new subtree is added to the roots of the evaluation poly-tree. All of these operations can be performed in time proportional to $ncLog(c)$, where $n$ is the number of nodes in the poly-tree and $c$ is the number of terms in any one agenda. However, note that on completion of these updates streams may not contain the same number of completed terms. The underlying theory has already been developed in [25]. The contribution here is simply to point out its applicability to incremental term computation.

**Model extension/reformulation**  We consider monotonic network growth only. Network extensions include both arc and node addition (we do not currently permit modifications to variable value spaces). Both addition of new nodes and addition of arcs to new nodes are trivial, neither affects the current evaluation polytree. Addition of arcs to existing nodes has two effects: first, it may create a new loop in the net, requiring that a marginalization be delayed. Second, it introduces a new variable (the new antecedent) at the point where the new arc has been introduced. Both of these consequences are handled similarly: existing terms in a stream must be split (conditioned) on the values of the new antecedent. This later is work in progress, and not fully implemented at this time.

## 6  Experimental Evaluation

We have been applying term computation to a variety of problems, but our core application is real-time decision-making [6] (although not discussed in this paper, the approach easily extends to arbitrary influence diagrams). Figure 4 shows how term computation using Z* search scales with problem size (number of components), as compared with exact, exhaustive evaluation using a traditional belief net inference algorithm (SPI, [25]). The two tasks are computation of the most likely composite hypothesis (MLCH), and computation of the optimal action over a range of alternatives including sensing and repair actions. The exact MLCH computation is performed using the algorithm by Li presented elsewhere in this conference. The decision evaluation requires MSEU estimation over a two stage decision problem (ie, we use one-step lookahead to estimate value of information for probe actions). Exact decision evaluation rapidly becomes intractable, while term computation scales more tractably. The problem is more difficult than it might seem: The decision network for the 4 component system contains 27 nodes (eleven in the first stage, eleven in the second stage, four outcome nodes, and the value node), many of which do not have the skewed property. Each component state node contains 4 values, includes an "unknown" behaviour mode with uniform distribution over outputs, and each input bit (1 for the one and two gate circuits, two for the four gate circuit, and 3 for the nine gate circuit, present and unobserved in the second decision stage) has uniform distribution over possible values it might take. Finally, the value node does not meet our definition for skewness.

| Gates | MLCH | | MSEU | |
|---|---|---|---|---|
| | TCS | Exact | TCS | Exact |
| 1 | .03/7 | .02/12 | 1.5/271 | .45/2308 |
| 2 | .04/20 | .04/64 | 3.7/607 | 3.7/30k |
| 4 | .11/33 | .8/128 | 4.6/902 | 900/5M |
| 9 | .29/83 | 24/183k | 45/4166 | ? |

Note: "n1/n2" indicates cpu-secs/#-of-terms created for TCS, cpu-sec/#-of-multiplications for Exact evaluation.

Figure 4: No fault

Several aspects of our approach are difficult to evaluate theoretically, and best examined experimentally. These include the use of Z*, marginalization, and substream caching.

| Gates | MLCH | | MSEU | |
|---|---|---|---|---|
| | TCS | Exact | TCS | Exact |
| 1 | .03/7 | .02/12 | .83/184 | .45/2308 |
| 2 | .19/40 | .04/64 | .19/40 | 3.7/30k |
| 4 | .40/104 | .8/128 | 7.4/1300 | 900/5M |
| 9 | 1.4/313 | 24/183k | 107/3113 | ? |

Note - all entries generated using Z* except the 9 component decision, which used greedy search.

Figure 5: One fault

First, why use Z*, a potentially exponential time method, for finding largest terms? In fact, while figure 4 was generated using Z* (except for the single fault 9 gate decision), in practice we often use a modified best-first strategy for decision evaluation[7]. Z* performs quite well for the MLCH task, handling the difficult nine gate case quite well. It has more trouble with decision evaluation. Its behavior is extremely sensitive to the quality of the factoring and the particular data available. With the best-first search, however, we have obtained decision times of 2-8 secs for both nominal and difficult scenarios. We are unsure at this time whether Z* is a practical control strategy for use in the kind of problem solving which has motivated this research, or more domain specific control mechanisms will be needed. Future research will be aimed at investigating this issue. When we abandon Z*, however, we lose theoretical guidance regarding how many terms to compute, and must rely on experience and heuristics.

Second, is marginalization worth it, under the assumption that only a few terms will be computed, and therefore marginalization opportunities will be rare? In fact, little marginalization occurs in typical applications to date. On the other hand, the overhead of checking for opportunities to marginalize is less than 10% of execution time.

Third, is caching of substreams worth it, for the same reason? Here the data is less ambiguous. The four component decision problem exceeds available space without substream caching.

## 7 Discussion

We have sketched a process which is essentially heuristic search for the set of bindings across a set of variables that maximizes the posterior probability across those variables. In another context, deKleer has referred to this as the "Most Likely Composite Hypothesis" problem [9], Henrion has described an algorithm for diagnosis in very large knowledge bases [15], Pearl has discussed the problem of "Distributed Revision of

---

[7]Details in extended technical report.

Composite Beliefs" [21], and Poole has sketched methods for probabilistically guided search [24]. Srinivas [27] treats a dual problem, that of obtaining the posterior probabilities of assumptions in an ATMS. From another perspective, Horvitz et al have been developing bounded conditioning as an approach to anytime probabilistic inference [16]. We believe the contributions of our work are several: (1) We have shown how this approach can be extended to arbitrary queries; (2) We have shown how, with caching and marginalization, an incremental probabilistic inference system based on computation of individual terms can be made as efficient at computing all terms (within a factor of $log(n)$) as the best algorithms for exact inference; (3) We have demonstrated that this process can be made incremental with respect to queries, evidence, and model revisions; (4) We have argued that such a system can serve as the basis for a tractable general-purpose low-level representation service.

Finally, a note regarding the relationship between this approach and propositional truth maintenance systems. Many of the internal dependency tracking mechanisms we have sketched are similar to those in an ATMS. There are several key differences. First, due to the loss of modularity in probabilistic inference [14], we propagate along the evaluation polytree rather than the original network. Second, The query driven nature of the control strategy permits us to marginalize over variables no longer needed on a path, avoiding the (potentially) exponential explosion of ATMS label size with network depth. Finally, ATMS nogood maintenance is replaced by Bayesian conditioning on evidence.

## 8 Conclusion

Problem solvers demand more interaction with an underlying representation service than is typically provided by current implementations either of truth maintenance or of efficient probabilistic inference in belief nets. We have sketched the current status of work in progress to develop an appropriate functional interface to a probabilistic representation service based on belief nets. This work is based on a redefinition of the basic inference task from exact computation of a prior or posterior probability distribution to computation of a single term, or conjunct, in that prior or posterior. It further provides incremental revision capabilities, rather than assuming a static network.